\documentclass[runningheads]{llncs}

\usepackage[T1]{fontenc}
\def\doi#1{\href{https://doi.org/\detokenize{#1}}{\url{https://doi.org/\detokenize{#1}}}}
\usepackage{graphicx}
\usepackage{listings}
\usepackage{amssymb} 
\usepackage{amsmath}

\usepackage{bm}

\usepackage{xcolor}

\usepackage{caption}
\usepackage{subcaption}

\usepackage[style=numeric,citestyle=numeric,backend=bibtex,backref=false,minnames=2,sorting=none]{biblatex}
\begin{document}
\title{Implicit Neural Representations for\\Generative Modeling of Living Cell Shapes}
\titlerunning{Generative Modeling of Living Cell Shapes}
%
\author{David~Wiesner\inst{1}\textsuperscript{*}
\and Julian~Suk\inst{2}
\and Sven~Dummer\inst{2}
\and \\ David~Svoboda\inst{1}
\and Jelmer M. Wolterink\inst{2}
}
%
\authorrunning{Wiesner et al.}
%
\institute{Centre for Biomedical Image Analysis, Masaryk University, Brno, Czech Republic
\and Department of Applied Mathematics \& Technical Medical Centre,\\ University of Twente, Enschede, The Netherlands\\[2ex]
\textsuperscript{*}Corresponding author: \email{wiesner@fi.muni.cz}}
\maketitle 
\begin{abstract}
Methods allowing the synthesis of realistic cell shapes could help generate training data sets to improve cell tracking and segmentation in biomedical images. Deep generative models for cell shape synthesis require a~light-weight and flexible representation of the cell shape. However, commonly used voxel-based representations are unsuitable for high-resolution shape synthesis, and polygon meshes have limitations when modeling topology changes such as cell growth or mitosis. In this work, we propose to use level sets of signed distance functions (SDFs) to represent cell shapes. We optimize a neural network as an implicit neural representation of the SDF value at any point in a 3D+time domain. The model is conditioned on a latent code, thus allowing the synthesis of new and unseen shape sequences. We validate our approach quantitatively and qualitatively on \textit{C. elegans} cells that grow and divide, and lung cancer cells with growing complex filopodial protrusions. Our results show that shape descriptors of synthetic cells resemble those of real cells, and that our model is able to generate topologically plausible sequences of complex cell shapes in 3D+time. 

\end{abstract}
\section{Introduction}
Deep learning has led to tremendous advances in segmentation and tracking of cells in high-resolution 2D and 3D spatiotemporal microscopy images~\cite{meijering2020bird}. To a large extent, this development has been driven by community projects like the Broad Bioimage Benchmark Collection~\cite{ljosa2012annotated} and Cell Tracking Challenge~\cite{ulman2017objective} that provide easily accessible biomedical image datasets. Nevertheless, the performance of deep learning methods is heavily dependent on the amount, quality, and diversity of the provided training data, and there is still a lack of diverse annotated datasets~\cite{dataavailability}. This fuels an interest in methods for synthesis of microscopy images and accompanying ground truth masks. 

The synthesis of microscopy images from ground truth masks has been widely studied~\cite{, Svoboda:MitoGen2017}, and has taken a major leap with the advent of generative adversarial networks~\cite{2019bohland, 2019bailo,2020bahr,2018fu, 2019baniukiewicz, 2019dunn, 2019han}. In this work, we focus on the synthesis of ground truth masks. Here, a key question is how cell shapes should be represented. A~range of parametric models have been proposed that use ellipses~\cite{2019bohland} or elliptical Fourier descriptors~\cite{2019scalbert} in 2D, statistical shape models~\cite{2020bahr} in 2D+time, ellipsoids~\cite{2019dunn, 2019han} (3D) and spherical harmonics~\cite{ducroz2012:spharmat} in 3D, or ellipsoids deformed using active contours in 3D+time~\cite{Svoboda:MitoGen2017}. Deep learning has led to the popularization of volumetric voxel-based representations in 3D due to their natural integration with CNN architectures~\cite{Wiesner:3dgan, 2018fu, 2019baniukiewicz}. However, the size of a 3D voxel mask, and thus its memory footprint, grows cubically with the shape resolution, making them unsuitable for complex cell shapes. Alternatives like polygonal meshes have limitations when modeling growth or mitosis in living cells~\cite{2016li}.  

\begin{figure}[t!]
\centering
\includegraphics[width=\textwidth]{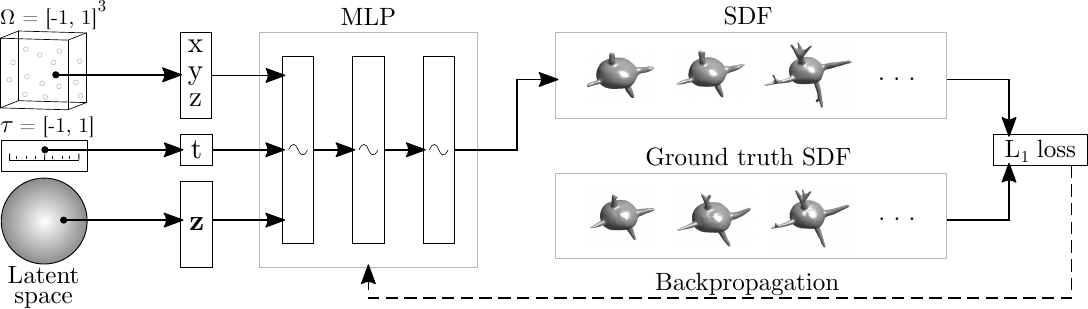}
\caption{Conceptual diagram of the proposed method. The multilayer perceptron (MLP) with \textit{sine} activation functions is given a latent code $\bm{z}$ sampled from a~multivariate normal distribution, coordinates $x, y, z$ from a~spatial domain $\mathrm{\Omega}$, and a temporal coordinate $t$ from temporal domain $\mathrm{\tau}$. The network is optimized to output the values of given spatio-temporal SDFs, whereas each latent code is optimized to describe a particular spatio-temporal SDF. The trained network is able to output SDF values based on a given latent code and space-time coordinates, where the optimized latent codes represent existing spatio-temporal SDFs from the training set. When given new latent codes, the trained network is able to infer new spatio-temporal SDFs and thus produce new evolving shapes.}
\label{fig:diagram}
\end{figure}

Here, we propose to model the cell surface as the zero level-set of a continuous signed distance function (SDF) in space and time. We represent this function via a multilayer perceptron (MLP) that takes as input spatial coordinates, a~time point, and a latent code. Following the DeepSDF model proposed by Park et al.~\cite{park2019deepsdf}, we jointly optimize this MLP and its latent space using a large set of 3D+time cell shape sequences. Once trained, the DeepSDF can synthesize completely new cell shape sequences at any spatial or temporal resolution (see Fig.~\ref{fig:diagram}). We demonstrate  on shapes of \textit{C. elegans}~\cite{2008murray} and lung cancer cells~\cite{sorokin2018filogen} that this single MLP provides topologically plausible cell shape, growth, and mitotic division. We show how periodic activations~\cite{sitzmann2020implicit} substantially improve the model's ability to represent highly complex shapes. Moreover, we show how the artificially generated shapes can be used as ground truth masks in benchmark datasets. To the best of our knowledge, this is the first generative model that implicitly represents cell shapes as differentiable and, therefore, trainable neural networks. 

\section{Method}
We propose to represent the evolution of a cell surface as the zero-level set of its evolving signed distance function (SDF). The SDF provides the Euclidean distance of any point in space to the nearest point on the cell surface at a point in time, where negative values are inside the surface and positive values outside. More precisely, let $\mathrm{\Omega} = [-1, 1]^3$ be a spatial domain, $\tau = [-1, 1]$ a temporal domain, and  $\mathcal{M}_t$ be a 2D manifold embedded in $\mathrm{\Omega}$ at time $t \in \tau$. For any point $\bm{x} = (x, y, z)\in \mathrm{\Omega}$, the $SDF_{\mathcal{M}_t}:\mathrm{\Omega} \rightarrow \mathbb{R}$ is defined as
\begin{eqnarray*}
    SDF_{\mathcal{M}_t}(\bm{x}) = \begin{cases}
\min_{\bm{u} \in \mathcal{M}_{t}} ||\bm{x} - \bm{u}||_2, \quad \quad \bm{x} \text{ outside } \mathcal{M}_{t} \\
0,  \qquad \qquad \qquad \qquad \quad \ \ \, \bm{x} \text{ belonging to } \mathcal{M}_{t} \\
-\min_{\bm{u} \in \mathcal{M}_{t}}||\bm{x} - \bm{u}||_2, \quad \bm{x} \text{ inside } \mathcal{M}_{t}
\end{cases}
\end{eqnarray*}
The zero-level set, and thus the surface of the cell at time $t$, is represented by all points where $SDF_{\mathcal{M}_t}(\cdot)=0$.

\subsection{Learning a Latent Space of Shapes}
Recent works have shown that the function $SDF_{\mathcal{M}_t}(\bm{x})$ can be approximated using a multi-layer perceptron (MLP) $f_\theta$ with trainable parameters $\theta$~\cite{sitzmann2020implicit,park2019deepsdf}. Such an MLP takes a coordinate vector $\bm{x}$ as input, and provides an approximation of $SDF_{\mathcal{M}_t}(\bm{x})$ as output. We here propose to condition the MLP on a time parameter $t \in \tau$ to provide an approximation of the SDF of $\mathcal{M}_t$ for arbitrary $t \in \tau$. In addition, the MLP can be conditioned on a latent space vector $\bm{z}$ drawn from a multivariate Gaussian distribution with a spherical covariance $\sigma^2 I$ (see Fig.~\ref{fig:diagram}). Combining these terms results in an MLP $f_\theta(\bm{x}, t, \bm{z})$ that approximates the SDF of the manifold $\mathcal{M}_t$ for arbitrary $t \in \tau$, given latent space vector $\bm{z}$. Here, we describe how we optimize such a model, or \textit{auto-decoder}~\cite{park2019deepsdf}, for cell shape sequences $\mathcal{M}_t$. 

We optimize the auto-decoder given a training set consisting of $N$ cell shape sequences. For each cell shape sequence, reference values of its SDF are known at a discrete set of points in $\mathrm{\Omega}$ and $\tau$. An important aspect of the auto-decoder model is that not only the parameters $\theta$ are optimized during training, but also the latent code $\bm{z}$ for each sequence. The loss function therefore consists of two components. The first component is the reconstruction loss that computes the $L_1$ distance between reference SDF values and their approximation by the MLP,~i.e.
\begin{eqnarray*}
\mathcal{L}_{recon}(f_\theta(\bm{x}, t, \bm{z}), SDF_{\mathcal{M}_t}(\bm{x})) = \|f_\theta(\bm{x}, t, \bm{z} ) - SDF_{\mathcal{M}_t}(\bm{x})\|_1,
\end{eqnarray*}
The second component is given by 
\begin{eqnarray*}
\mathcal{L}_{code}(\bm{z}, \sigma) = \frac{1}{\sigma^2} \|\bm{z}\|^{2}_{2}.
\end{eqnarray*}
This term, with regularization constant $\frac{1}{\sigma^2}$, ensures that a compact latent space is learned and improves the speed of convergence~\cite{park2019deepsdf}. Note that $\sigma$ in this term corresponds to the Gaussian distribution of the latent vectors, and that latent vector $\bm{z}$ is fixed for one cell shape sequence. During training of the auto-decoder we have access to a training set of $N$ cell shape sequences from which we extract mini-batches of points, and thus the full loss function becomes
\begin{eqnarray*}\label{network}
\mathcal{L}(\theta, \{\bm{z}_i\}_{i=1}^N, \sigma) = \mbox{\Large\( \mathbb{E}_{(\bm{x}, t)} \)}\left(\sum_{i=1}^N \mathcal{L}_{recon}(f_\theta(\bm{x}, t, \bm{z}_i), SDF_{\mathcal{M}_t^i}(\bm{x})) +\mathcal{L}_{code}(\bm{z}_i, \sigma)\right),
\end{eqnarray*}
where each sequence $\{\mathcal{M}_t^i\}_{i = 1}^N$ is assigned a latent code $\bm{z}_i$.

\subsection{Neural Network Architecture}
The function $f_\theta(\bm{x}, t, \bm{z}_i)$ is represented by an MLP. In all experiments, we used a~network with 9 hidden layers, each containing 128 units. We present experiments with two activation functions. First, the commonly employed rectified linear unit (ReLU) $\sigma(x) = \max(0,x)$. While this activation function produces good results on low-frequency signals, an accurate representation of high-frequency information is not possible due to its inherent low frequency bias~\cite{mildenhall2020nerf}. Therefore, we also present results with periodic activation functions $\sigma(x) = \sin(x)$ (\textit{sine}). The weights of layers using \textit{sine} activations have to be initialized by setting a~parameter $\omega$ to ensure a good convergence~\cite{sitzmann2020implicit}. The value of $\omega$ directly affects the range of frequencies that the model is able to represent, where low values of $\omega$ encourage low frequencies and smooth surfaces, and high values favor high-frequencies and finely detailed surfaces.
Initial latent code vectors $\bm{z}_i$ of size 192 were sampled from $\mathcal{N}(0,0.1^2)$ and inserted in the first, fifth, and eight layer of the network to improve reconstruction accuracy~\cite{park2019deepsdf}. Inserting this information into more layers did not yield measurable improvement. Moreover, the coordinates $\bm{x}$ and $t$ were given to all hidden layers. In this case, we found out that the model would not converge on long spatio-temporal sequences without this additional inserted information.

\subsection{Data} \label{data}
To demonstrate modeling of different phenomena occurring during the cell cycle, we selected two existing annotated datasets for our experiments. First, a~population of real \textit{C. elegans} developing embryo cells that grow and divide~\cite{2008murray} ($\mathcal{D}_\text{cel}$). Second, synthetic actin-stained A549 lung adenocarcinoma cancer cells with growing filopodial protrusions~\cite{sorokin2018filogen} ($\mathcal{D}_\text{fil}$). Both datasets are produced in fluorescence microscopy modality and contain full 3D time-lapse annotations, i.e., they consist of pairs of microscopy images and corresponding segmentation masks. The \textit{C. elegans} dataset has a resolution of $708\times512\times35$ voxels (voxel size: $90\times90\times1000$ nanometers) and was acquired with a one-minute time step, whereas the filopodial cell dataset has a resolution of $300\times300\times300$ voxels (voxel size: $125\times125\times125$ nanometers) and was simulated with a 20-second time step.

The data preparation and visualization algorithms were implemented in Matlab R2021a.
Using the segmentation masks of evolving shapes from $\mathcal{D}_\text{cel}$ and $\mathcal{D}_\text{fil}$, we prepared 66 diverse 3D+time SDF sequences, 33 for \textit{C. elegans} and 33 for filopodial cells, with each sequence having shapes at 30 time points. As the cell surfaces occupy only a fraction of the considered 3D space, we sampled 70\% of SDF points around the cell surface and remaining 30\% in the rest of the space~\cite{park2019deepsdf}. This allowed us to save memory in comparison to uniform sampling that would use considerable amount of data points to represent empty space around the object. We will refer to these SDF datasets as $\mathcal{D}_\text{cel}^{SDF}$ and $\mathcal{D}_\text{fil}^{SDF}$.

\section{Experiments and Results}
The auto-decoder network was implemented in Python using PyTorch.
We trained separate models on $\mathcal{D}_\text{cel}^{SDF}$ and $\mathcal{D}_\text{fil}^{SDF}$ for 1250 epochs. The weights were optimized using Adam with a learning rate $10^{-4}$ that decreases every 100 epochs by multiplication with a factor $0.5$. Models were trained both with ReLU activation functions and periodic activation functions. For model utilizing the \textit{sine} activation functions, the $\omega$ parameter for weight initialization was set to 30. Because the trained model is continuous, it can be used to generate point clouds, meshes, or voxel volumes~\cite{park2019deepsdf}. In this work, we use voxel volumes for quantitative evaluation and meshes obtained using marching cubes~\cite{lorensen1987marching} for visualization. With a trained model, generating one 30-frame 3D+t SDF sequence with $30\times256\times256\times256$ uniform samples took 40 seconds on an NVIDIA A100 and required 3 GB of GPU memory. We used these models to perform a series of experiments.

\begin{figure}[t!]
\centering
\includegraphics[width=\textwidth]{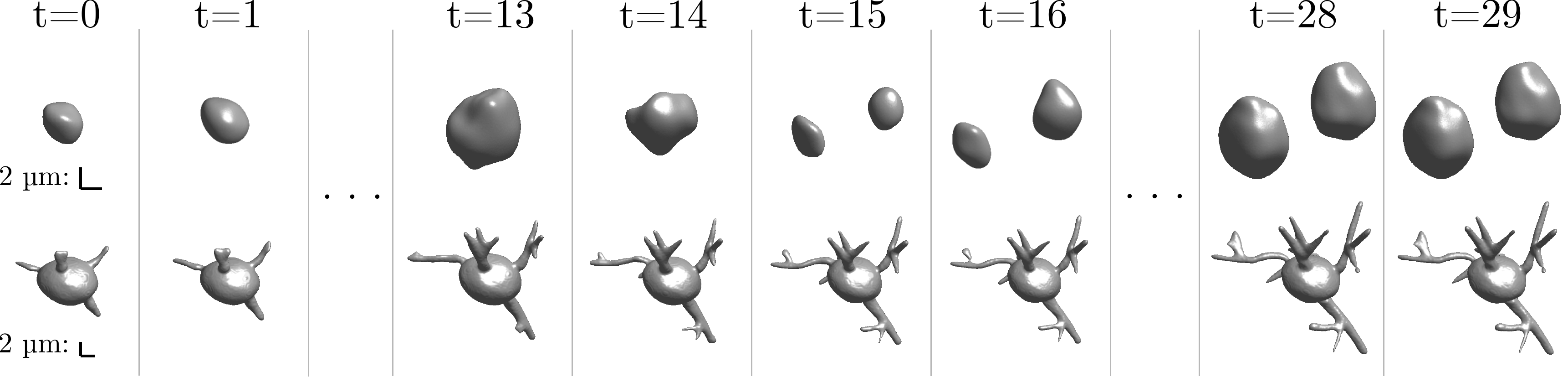}
\caption{New spatio-temporal sequences produced using the proposed method. The figure shows 3D renderings of cell surface meshes of \textit{C. elegans} (top row) and lung cancer cell with filopodial protrusions (bottom row) at selected time points. Each sequence has 30 frames, with mesh surfaces obtained from the inferred 3D+t SDFs having $30\times256\times256\times256$ samples.}
\label{fig:newsequences}
\end{figure}

\subsection{Reconstruction of Cell Sequences}
\label{reconstruction}
In our first experiment, we evaluated the ability of the trained model to reconstruct a cell shape sequence in the training set, given its latent code $\bm{z}$. To evaluate the reconstruction accuracy, we compute the Jaccard index (JI) on $256\times256\times256$ voxel volumes obtained from training ($\mathcal{D}_\text{cel}^{SDF}$ and $\mathcal{D}_\text{fil}^{SDF}$) and reconstructed SDF sequences. JI is computed on individual frames, i.e., we obtain one value for each pair of training and reconstructed frame. We computed mean and standard deviation to quantify the similarity of the reconstructed datasets. For \textit{C. elegans}, we obtained JI: $0.782\pm0.033$ (ReLU model) and $0.821\pm0.029$ (\textit{sine} model), for filopodial cells: $0.833\pm0.051$ (ReLU model) and $0.880\pm0.022$ (\textit{sine} model). The JI values show that the reconstruction accuracy of the auto-decoder using \textit{sine} activation functions is measurably higher. Visually, the thin protrusions of the filopodial cells lost their sharp edges and occasionally merged together with the ReLU model. In the remainder of this work, we present results for models using \textit{sine} activation functions.

\begin{figure}[t!]
\centering
\begin{subfigure}[b]{0.327\textwidth}
    \centering
    \includegraphics[width=\textwidth]{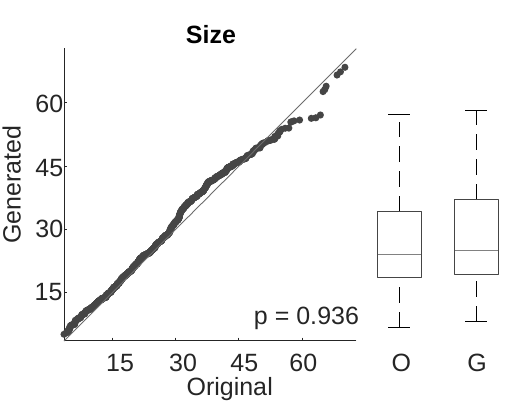}
\end{subfigure}
\hfill
\begin{subfigure}[b]{0.327\textwidth}
    \centering
    \includegraphics[width=\textwidth]{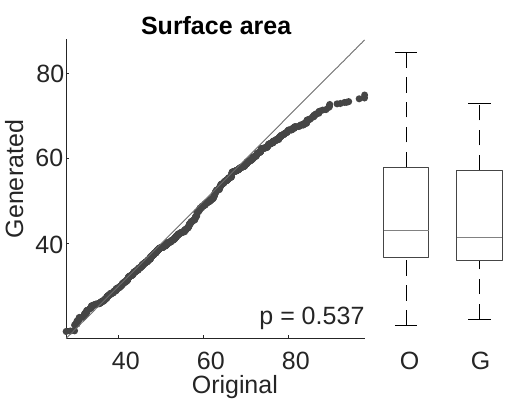}
\end{subfigure}
\hfill
\begin{subfigure}[b]{0.327\textwidth}
    \centering
    \includegraphics[width=\textwidth]{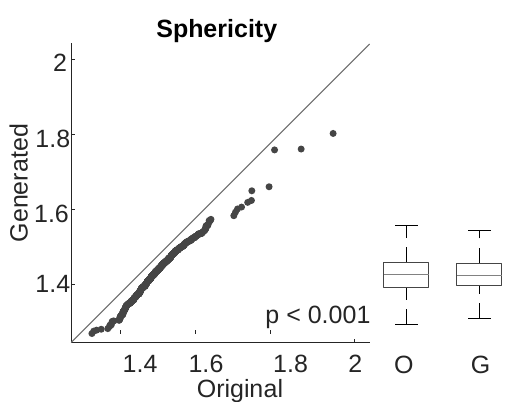}
\end{subfigure}

\begin{subfigure}[b]{0.327\textwidth}
    \centering
    \includegraphics[width=\textwidth]{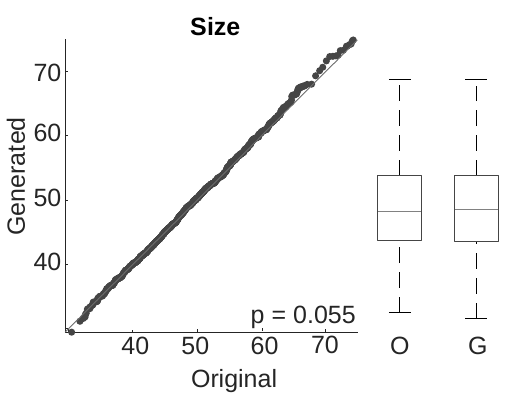}
\end{subfigure}
\hfill
\begin{subfigure}[b]{0.327\textwidth}
    \centering
    \includegraphics[width=\textwidth]{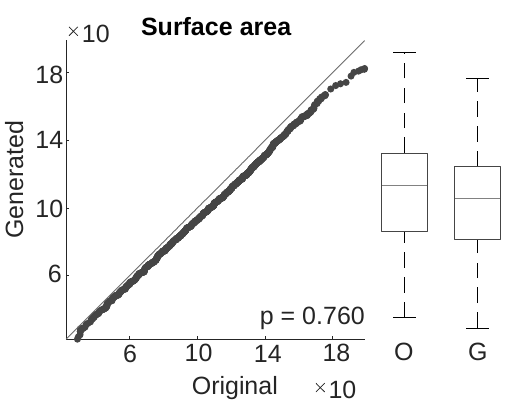}
\end{subfigure}
\hfill
\begin{subfigure}[b]{0.327\textwidth}
    \centering
    \includegraphics[width=\textwidth]{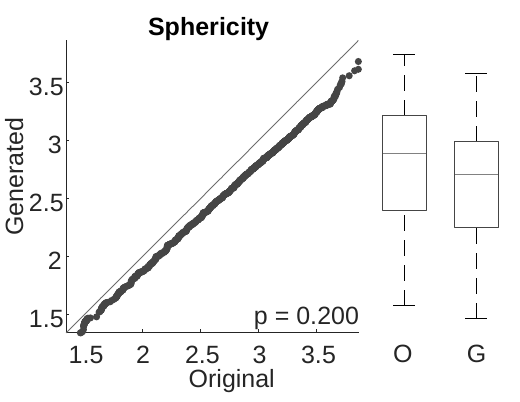}
\end{subfigure}
\caption{Quantile-quantile (QQ) plots and boxplots for the shape descriptor distributions of original (O) and new generated (G) shapes of C. elegans (top row) and filopodial cells (bottom row). The QQ plots show the ideal identical distribution as a straight line and include the p-values of the Kolmogorov-Smirnov test. The whiskers of the boxplots correspond to the min and max values, the boxes represent the interquartile ranges, and the thin marks show the median values.}
\label{fig:descriptors}
\end{figure}

\begin{table}[t!]
\caption{Shape descriptors (mean $\pm$ standard deviation) of original (O) and new shapes generated (G) using the proposed method.
}
\label{descriptors}
\centering
\begingroup
\setlength{\tabcolsep}{7pt}
\renewcommand{\arraystretch}{1.1}
\resizebox{\textwidth}{!}{
\begin{tabular}{|c|c|c|c|}
\hline
    \textbf{Dataset} &  \textbf{Size} [${\mu}m^3$] & \textbf{Surface area} [${\mu}m^2$] & \textbf{Sphericity} \\
\hline
    \textit{C. elegans} (O) &  $34.7\pm12.5$ & ${\color{white}0}62.8\pm16.1$ & $1.54\pm0.09$ \\
\hline
    \textit{C. elegans} (G) &  $36.5\pm11.5$ & ${\color{white}0}65.2\pm14.6$ & $1.54\pm0.06$ \\
\hline
    \textit{Filopodial cells} (O) &  $50.8\pm3.1{\color{white}0}$ & $117.8\pm19.9$ & $2.64\pm0.51$ \\
\hline
    \textit{Filopodial cells} (G) &  $51.7\pm3.2{\color{white}0}$ & $115.0\pm19.5$ & $2.50\pm0.49$ \\
\hline
\end{tabular}
}
\endgroup
\end{table}

\subsection{Generating New Cell Sequences}
In our second experiment, we used the trained auto-decoders to produce new cell shape sequences (see Fig.~\ref{fig:newsequences}). For the \textit{C. elegans} cells, we randomly generated 33 new latent codes $\bm{z}$ by sampling from $\mathcal{N}(0,0.1^2)$. For filopodial cells, a noise vector sampled from $\mathcal{N}(0,0.01^2)$ was added to the 33 learned latent vectors for $\mathcal{D}_\text{fil}^{SDF}$ reconstruction. The produced latent codes were given to the trained auto-decoders to produce 33 new \textit{C. elegans} sequences and 33 new filopodial sequences. To investigate how realistic the produced shapes are, we evaluated the similarity between the distribution of real sequences and the distribution of generated sequences using shape descriptors computed on their voxel representations, i.e., cell size (volume) in ${\mu}m^3$, surface area in ${\mu}m^2$, and sphericity~\cite{dipimage}. We compared the descriptor distributions using quantile-quantile plots, box plots, and the Kolmogorov-Smirnov (KS) test (see Fig.~\ref{fig:descriptors}). The plots show that the new shapes exhibit high similarity to the ones from the training sets. The KS test retained the null hypothesis ($p > 0.05$) that the descriptors are from the same distribution at $5\%$ significance level for all tests except for the sphericity of the generated \textit{C. elegans} cells, which exhibits a modest shift to lower values. For the mean and standard deviation values of each descriptor, see Table~\ref{descriptors}.

\subsection{Temporal Interpolation}
Because the 3D+t SDF representation is continuous, we can use the trained auto-decoder to produce sequences in arbitrary spatial and temporal resolution without the need of additional training. In particular, the temporal interpolation can be used to improve segmentation results~\cite{2014coca}. To evaluate the interpolation accuracy, we trained the auto-decoder on 33 filopodial sequences, but with half the number of frames (15) in each sequence. We then used this model to reconstruct the sequences with a double framerate (30). Similarly to Sec.~\ref{reconstruction}, we compared the reconstruction accuracy using JI with respect to the $\mathcal{D}_\text{fil}^{SDF}$ dataset. We obtained a JI of $0.868\pm0.044$, which is very close to the accuracy of the model trained on the ``full'' $\mathcal{D}_\text{fil}^{SDF}$ dataset ($0.880\pm0.022$).

\begin{figure}[t!]
\centering
\begin{subfigure}[b]{0.24\textwidth}
    \centering
    \includegraphics[width=\textwidth]{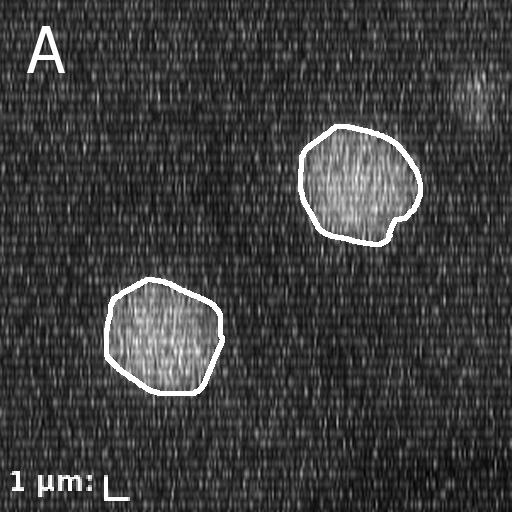}
\end{subfigure}
\hfill
\begin{subfigure}[b]{0.24\textwidth}
    \centering
    \includegraphics[width=\textwidth]{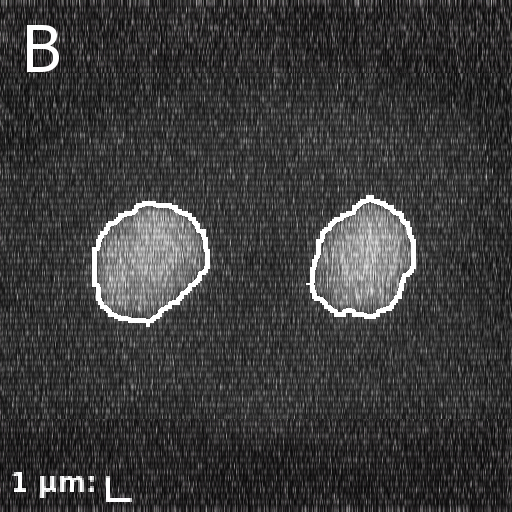}
\end{subfigure}
\hfill
\begin{subfigure}[b]{0.24\textwidth}
    \centering
    \includegraphics[width=\textwidth]{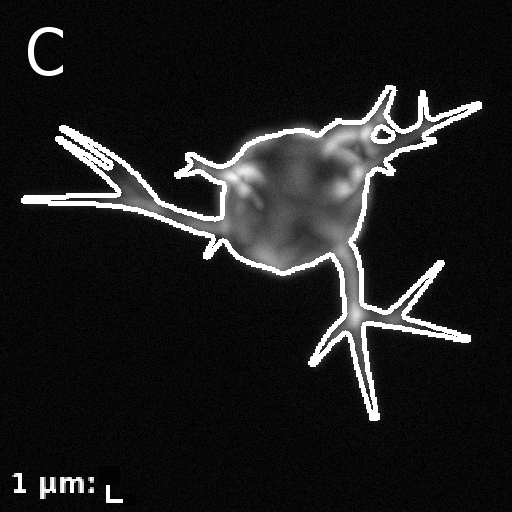}
\end{subfigure}
\hfill
\begin{subfigure}[b]{0.24\textwidth}
    \centering
    \includegraphics[width=\textwidth]{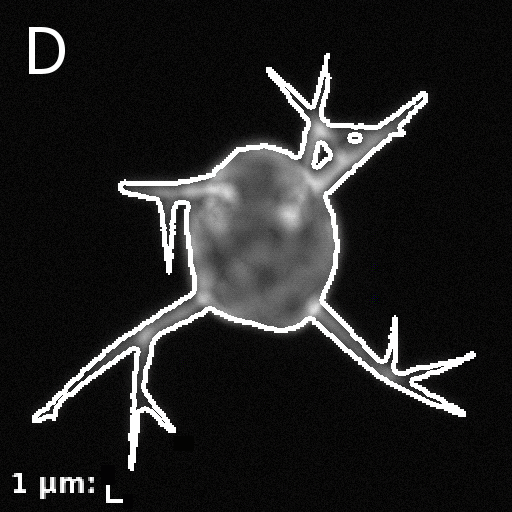}
\end{subfigure}
\caption{Comparison of original (A, C) and generated (B, D) images of \textit{C. elegans} (A, B) and filopodial cells (C, D). The images are maximum intensity projections of one frame from the respective time-lapse datasets. The respective segmentation masks are represented as white contours. The masks were obtained using the proposed method and the texture was produced using a conditional GAN.}
\label{fig:texture}
\end{figure}

\subsection{Generating Benchmarking Datasets}
To demonstrate the application of the method, we produced a new 2D+time benchmarking dataset of \textit{C. elegans} cells that grow and divide. Furthermore, we produced a new 2D+time dataset of cells with growing filopodial protrusions. Both datasets contain pairs of textured cell images and ground truth segmentations. The segmentation masks are maximum intensity projections of the voxel volumes of SDFs produced using the proposed method. The texture was generated using a conditional GAN, more specifically pix2pixHD~\cite{2018wang}, which was trained on maximum intensity projections of textures and corresponding masks from the $\mathcal{D}_\text{cel}$ and $\mathcal{D}_\text{fil}$ datasets. For a visual comparison, see Fig.~\ref{fig:texture}.

\section{Discussion and Conclusion}
We have presented a deep learning-based framework that can be used for accurate spatio-temporal representation and synthesis of highly-detailed evolving shapes and structures in microscopy imaging. To achieve this, we use a fully connected neural network to learn implicit neural representations of spatio-temporal SDFs. Owing to the employed periodic activation functions and SDF data representation, the method allows for shape synthesis with virtually unlimited spatial and temporal resolution at an unprecedented level of detail that would not have been possible with existing voxel-based methods~\cite{Svoboda:MitoGen2017, 2018fu, 2019baniukiewicz} and models utilizing ReLU activations~\cite{park2019deepsdf, remelli2020meshsdf}. The produced SDFs can be converted to mesh-based, voxel-based, or point cloud representations, depending on the desired application. The proposed model is simple and can be easily trained on a common workstation to produce a desired class of shapes without the need for laborious customization or an expensive computational platform. 

We presented the results and quantitative evaluation on two diverse datasets, the \textit{C. elegans} embryo cells and the actin-stained A549 lung adenocarcinoma cancer cells with protrusions. The modeling of cell growth and mitosis facilitates gaining better understanding of cell development and can be used for deriving accurate quantitative models, e.g., for embryogenesis. Specifically, the A549 lung cancer cells are subject to an active research because filopodia and their relationship to cell migration is of great importance to understanding of wound healing, embryonic development, or the formation of cancer metastases. For these two cell classes, we produced benchmarking datasets for training and evaluation of image analysis algorithms.

The framework can be used for increasing spatial and temporal resolution of existing datasets, for data augmentation, or for generating brand-new benchmarking datasets. We have here presented the application of this generative model for cell shape synthesis, but organisms present many spatiotemporal dynamics at micro- and macro scales. The model could, for example, be extended to synthesize brain atrophy in patients with Alzheimer's disease, or the progression of abdominal aortic aneurysms.

In conclusion, conditional implicit neural representations or auto-decoders are a feasible representation for generative modeling of living cells. 

\subsubsection{Note} The source codes, models, and datasets are made publicly available at: \url{https://cbia.fi.muni.cz/research/simulations/implicit_shapes.html}

\subsubsection{Acknowledgements} This work was partially funded by the 4TU Precision Medicine programme supported by High Tech for a Sustainable Future, a framework commissioned by the four Universities of Technology of the Netherlands. Jelmer M. Wolterink was supported by the NWO domain Applied and Engineering Sciences VENI grant (18192). David Wiesner was supported by the Grant Agency of Masaryk University under the grant number MUNI/G/1446/2018. David Svoboda was supported by the MEYS CR (Projects LM2018129 and
CZ.02.1.01/0.0/0.0/18\_046/0016045 Czech-BioImaging). 

This preprint has not undergone peer review (when applicable) or any post-submission improvements or corrections. The Version of Record of this contribution is published in ``Medical Image Computing and Computer Assisted Intervention (MICCAI) 2022 Proceedings, Part IV (LNCS 13434)'', and is available online at: \url{https://dx.doi.org/10.1007/978-3-031-16440-8_6}
%
%
%
%
\printbibliography
\end{document}